\documentclass[letterpaper]{article} 
\usepackage{aaai25}  
\usepackage{times}  
\usepackage{helvet}  
\usepackage{courier}  
\usepackage[hyphens]{url}  
\usepackage{graphicx} 
\usepackage{natbib}  
\usepackage{caption} 
\usepackage{algorithm}
\usepackage{algorithmic}

\usepackage{newfloat}
\usepackage{listings}
\DeclareCaptionStyle{ruled}{labelfont=normalfont,labelsep=colon,strut=off} 
\floatstyle{ruled}
\newfloat{listing}{tb}{lst}{}
\floatname{listing}{Listing}
\usepackage{subcaption}
\usepackage{multirow}
\usepackage{amssymb}
\usepackage{amsmath}

\title{SAUGE: Taming SAM for Uncertainty-Aligned Multi-Granularity Edge Detection}
\author{
    Xing Liufu\textsuperscript{\rm 1},
    Chaolei Tan\textsuperscript{\rm 1},
    Xiaotong Lin\textsuperscript{\rm 1},
    Yonggang Qi\textsuperscript{\rm 2},
    Jinxuan Li\textsuperscript{\rm 1},
    Jian-Fang Hu\textsuperscript{\rm 1,3,4}\thanks{Corresponding author.}
}
\affiliations{
    \textsuperscript{\rm 1}School of Computer Science and Engineering, Sun Yat-sen University, Guangzhou, China\\
    \textsuperscript{\rm 2}Beijing University of Posts and Telecommunications, Beijing, China\\
    \textsuperscript{\rm 3}Key Laboratory of Machine Intelligence and Advanced Computing, Ministry of Education, Guangzhou, China\\
    \textsuperscript{\rm 4} Guangdong Province Key Laboratory of Information Security Technology, China\\
    \{liufux,tanchlei,linxt29,lijx267\}@mail2.sysu.edu.cn, qiyg@bupt.edu.cn, hujf5@mail.sysu.edu.cn
}

\begin{document}

\maketitle

\begin{abstract}
Edge labels are typically at various granularity levels owing to the varying preferences of annotators, thus handling the subjectivity of per-pixel labels has been a focal point for edge detection. Previous methods often employ a simple voting strategy to diminish such label uncertainty or impose a strong assumption of labels with a pre-defined distribution, e.g., Gaussian. In this work, we unveil that the segment anything model (SAM) provides strong prior knowledge to model the uncertainty in edge labels. Our key insight is that the intermediate SAM features inherently correspond to object edges at various granularities, which reflects different edge options due to uncertainty. Therefore, we attempt to align uncertainty with granularity by regressing intermediate SAM features from different layers to object edges at multi-granularity levels. In doing so, the model can fully and explicitly explore diverse ``uncertainties'' in a data-driven fashion. Specifically, we inject a lightweight module ($\sim 1.5\%$ additional parameters) into the frozen SAM to progressively fuse and adapt its intermediate features to estimate edges from coarse to fine. It is crucial to normalize the granularity level of human edge labels to match their innate uncertainty. For this, we simply perform linear blending to the real edge labels at hand to create pseudo labels with varying granularities. Consequently, our uncertainty-aligned edge detector can flexibly produce edges at any desired granularity (including an optimal one). Thanks to SAM, our model uniquely demonstrates strong generalizability for cross-dataset edge detection. Extensive experimental results on BSDS500, Muticue and NYUDv2 validate our model's superiority.
\end{abstract}

%

\section{Introduction}

\begin{figure}[htbp]
	\centering
	\includegraphics[width=0.95\linewidth]{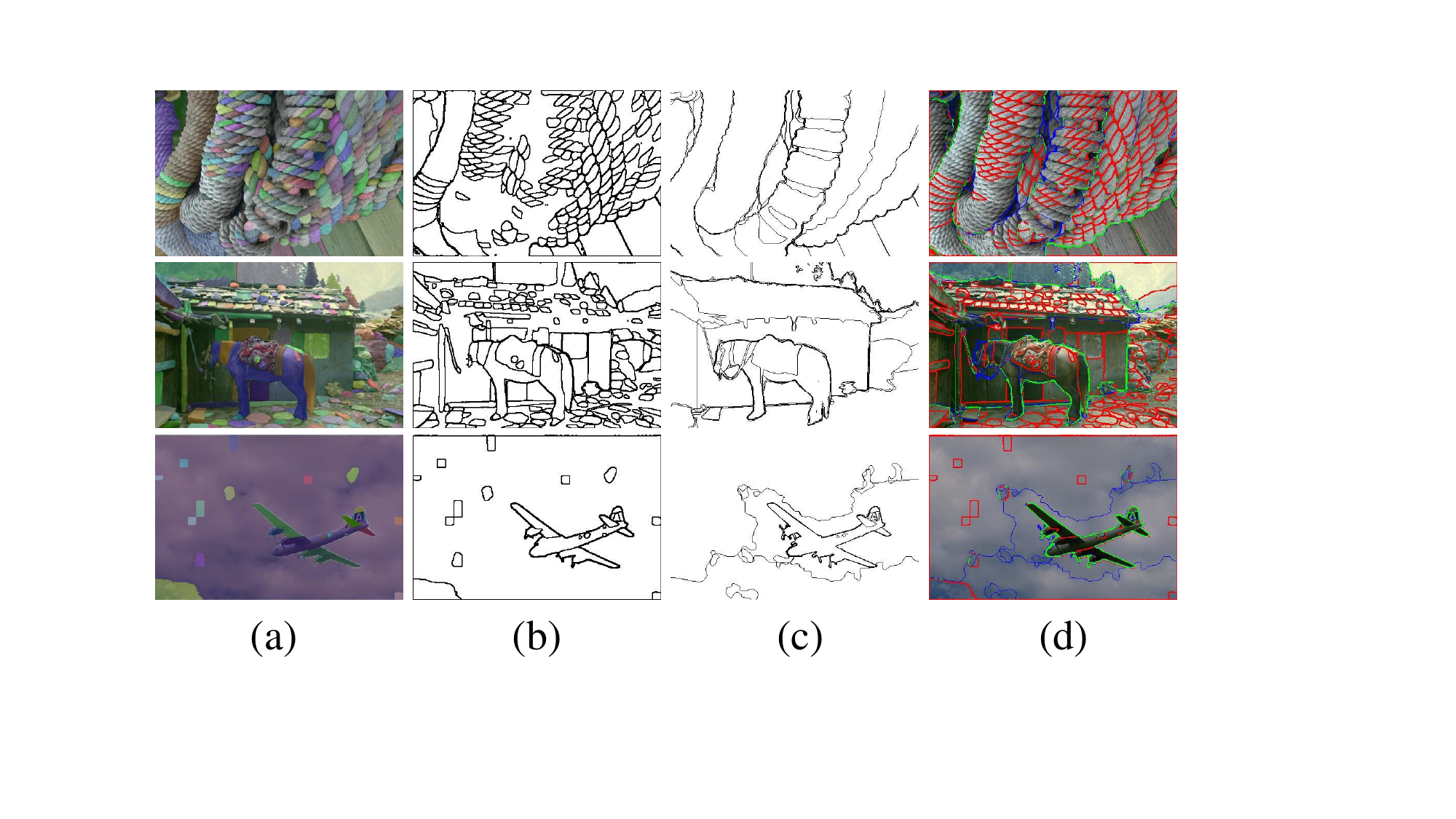}
	\caption{Comparison of edges obtained by SAM and manually annotated ground truth: (a) shows images from BSDS set; (b) illustrates edge maps generated by SAM; (c) presents edge annotations; (d) compares (b) and (c), where green indicates the shared edges, red represents edges in (b) but not in (c), and blue represents edges in (c) but not in (b).}
	\label{fig:sam_guide}
\end{figure}

Edge detection is a fundamental AI task in low-level vision that plays a crucial role in image understanding. 
It is of great value in supporting various high-level vision tasks, including semantic segmentation~\cite{yu2021coupled}, image enhancement~\cite{nazeri2019edgeconnect, xu2023low}, object detection~\cite{qin2019basnet,liu2020dynamic,yao2023object}, image translation~\cite{jiang2023masked}, etc. 

However, due to the observers’ diverse levels of visual perception \cite{zhou2023treasure}, there are often multiple ground truth edge maps for a given image. The inconsistency between different edge maps can cause uncertainty in decision-making, complicate the model training, and reduce the overall performance.
The scarcity of human annotations may further exacerbate this issue.

Most previous works \cite{liu2017richer,he2019bi,pu2022edter} tend to overlook the uncertainty in annotations by treating each edge map as equally valid, or fuse variant edge maps into a unified version by a simple voting strategy. As a result, these methods typically fail to capture the inherent subjectivity in the annotations. They can only produce a single edge map for a given image, lacking the ability to control granularity. This limitation severely declines their applicability and scalability in real-world scenarios.

To cope with the uncertainty issue of edge labels is non-trivial and remains relatively under-explored. UAED \cite{zhou2023treasure} handles edge labels from a probabilistic perspective. It is assumed that the uncertainty of edge labels could be modeled using a Gaussian distribution, allowing the generation of multiple edges by sampling from the distribution. However, the distribution assumption imposes strict constraints, yielding limited diversity and a lack of control over granularity. RankED \cite{cetinkaya2024ranked} approached to calculate the label certainty of each pixel among different annotators, then favor the pixels with higher confidence during training. Consequently, pixel annotations with high uncertainty are largely overlooked.
More recently, MuGE \cite{zhou2024muge} devised a granularity-controllable edge detector, which remedies label uncertainty by producing edge maps at various granularity levels.
They naively assign a binary granularity score to its most simple (0) and complex (1) edge maps for an image to train a binary classifier, which is then applied to assign granularity scores (0-1) to rest edge maps. Then the granularity score is explicitly embedded in the edge detector, making it measurable and controllable. Unfortunately, this may easily introduce bias as a complex edge map might be incorrectly marked as simple if all the annotations are complex, and vice versa.

In this work, following the core idea of MuGE which models uncertainty by generating multi-scale edge maps, we seek to develop an edge detector that can produce granularity-controllable outputs without requiring granularity labels for edge maps. This can be achieved by progressively fusing and projecting intermediate features of the segmentation foundation model SAM~\cite{kirillov2023segment} to edge map variants with increasing granularities.
We demonstrate that SAM is highly effective in edge detection in the multi-scale setting. Intuitively, SAM excels in locating object boundaries, thus potentially providing strong prior knowledge of edge maps. Besides, we find out that the features from intermediate layers of SAM naturally encode rich information about various granularities for object edges.

However, unlocking the full potential of SAM for multi-granularity edge detection presents significant challenges. As aforementioned, SAM inclines to provide redundant details and coarse object boundaries rather than detailed internal edges, as shown in Figure \ref{fig:sam_guide} (d). To fill the gap and capture uncertainty through granularity modeling, we incorporate a lightweight feature transfer network into SAM that intermediate features are gradually fused and projected into side outputs of edges at increasingly complex levels. 
To supervise the transfer network, linear blending is performed to the real edge labels to synthesize pseudo edges at diverse granularities. This step crucially normalizes the granularity levels of all edge labels, easing the convergence of network training.
Moreover, we develop a novel diversity loss to encourage the obtained side edge maps to be sufficiently diverse, further enhancing uncertainty modeling.

Our main contributions can be summarized as follows:
\begin{itemize}
\item We propose a novel edge detector, named SAUGE, which sidesteps the uncertainty difficulty by modeling edge detection in a multi-granularity setting based on SAM. The obtained edge detector enables edge detection at any desired granularity level.
\item A lightweight module is proposed to be injected into the frozen SAM to progressively fuse and transfer its intermediate features to generate edges from coarse to fine. Linear blending is performed to real edge labels to synthesize normalized supervision cues.
\item We conduct extensive experiments on the BSDS500, Multicue and NYUDv2 datasets, and the results demonstrate that our model achieves the new state-of-the-art and suggest a strong generalizability on unseen datasets.
\end{itemize}

\section{Related Work}
\begin{figure*}[htbp]
	\centering
	\includegraphics[width=0.9\linewidth, height=0.433\linewidth]{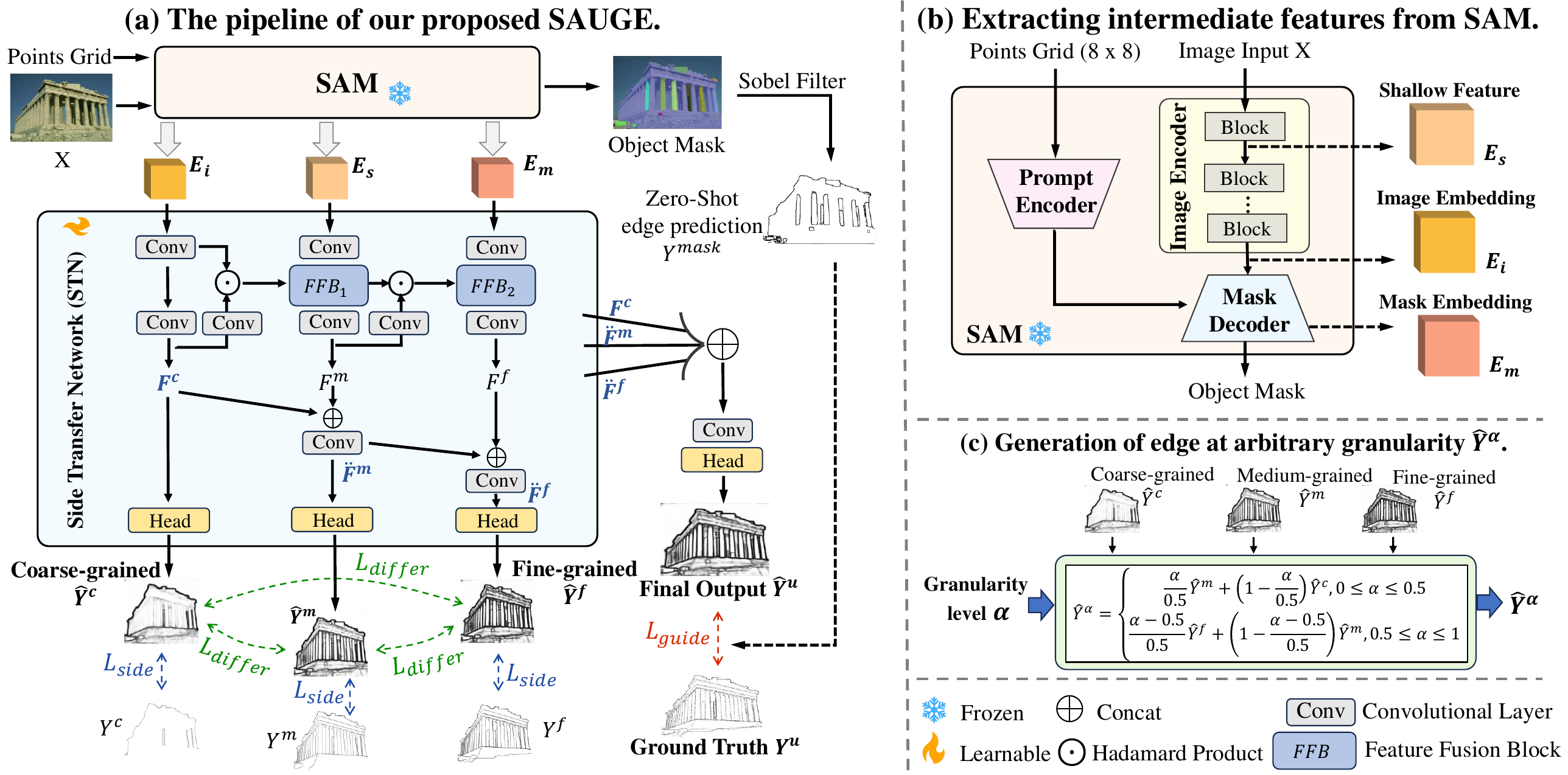}
	\caption{The overall framework of SAUGE. (a) illustrates the pipeline of SAUGE. 
    We extract the intermediate SAM features and feed them into STN, which constructs edges at multiple granularity levels to align uncertainty with granularity. 
    The final output $\hat{Y}^u$ is obtained by merging the features of side outputs. We devise losses $\{L_{side}, L_{differ}, L_{guide}\}$ to supervise the side outputs, promote the pairwise diversity among them, and guide edge learning using SAM masks, respectively. (b) shows the features extracted from SAM. (c) demonstrates the generation of $\hat{Y}^\alpha$ at any granularity level $\alpha$ in a controllable manner.}
	\label{fig:model}
\end{figure*}
 
\subsubsection{Edge detection.} Edge detection has been a significant research area for a long time. Traditional techniques such as~\cite{kittler1983accuracy,canny1986computational} rely on gradient computation on low-level features which are easy to be affected by noise. 
Over the past decade, deep learning-based approaches, such as \cite{shen2015deepcontour,liu2016learning,he2019bi,su2021pixel}, have risen to prominence, focusing on designing network architectures that surpass human-level performance. The majority of these methods, including RCF~\cite{liu2017richer}, leverage pre-trained VGG16~\cite{simonyan2014very} as their backbone. More recently, advanced methods like EDTER~\cite{pu2022edter} incorporated Transformers to enhance edge detection.
Research based on uncertainty explores the uncertainty caused by multiple labels. UAED~\cite{zhou2023treasure} models edge maps as multivariate Gaussian distribution and use predicted variance to measures the uncertainty, RankED~\cite{cetinkaya2024ranked} sorts pixels to balance edge and non edge pixels and promote higher label certainty for high confidence edges. MuGE~\cite{zhou2024muge} integrates encoded edge granularity into feature maps to produce edge maps at multiple granularities for alleviating uncertainty. Recent image generation methods such as DiffusionEdge~\cite{ye2024diffusionedge} applies diffusion models to generate crisp edge maps.

Most of these methods do not utilize prior knowledge from advanced tasks. In contrast, we explore the priors of semantic-aware features in SAM and construct multiple granularity edge maps to address the uncertainty. 

\subsubsection{Exploring SAM for downstream tasks.}
The Segment Anything Model (SAM)~\cite{kirillov2023segment} accepts intuitive prompts (points or bounding boxes), and has established a new benchmark in natural image segmentation. It has demonstrated impressive performance across various downstream tasks, including medical imaging~\cite{ma2024segment, gu2024boosting}, object segmentation in challenging conditions~\cite{chen2024robustsam,zhang2024irsam,ke2024segment} and image inpainting~\cite{yu2023inpaint}.

The work most closely related to ours is EdgeSAM~\cite{yang2024boosting}, which introduces an adapter module to fine-tune SAM for edge detection. However, their approach uses SAM in a simplistic manner, without fully exploring its potential and modeling the uncertainty. In contrast, we leverage the granularity-aware prior knowledge embedded in SAM's features and attempt to align uncertainty with granularity.

\section{Method}
The overall framework of the proposed uncertainty-aligned edge detector (SAUGE) is presented in Figure \ref{fig:model}. As shown, our SAUGE is built upon the pre-trained SAM, utilizing the proposed lightweight Side Transfer Network (STN) to explicitly explore the knowledge embedded in SAM for multiple granularity edge detection. STN gradually uses Feature Fuse Block (FFB) to fuse intermediate features extracted from different stages of SAM to supplement edge-aware details, thereby gradually constructing side outputs representing edge at varied granularity levels. Finally, STN fuses all the features of side outputs to generate final output. To ensure the quality of side outputs, we construct pseudo labels corresponding to varied granularity levels to supervise and emphasize diversity of side outputs. We also propose to use the masks outputted by SAM to improve our edge learning.

\subsection{Aligning uncertainty through granularity modeling}
The proposed Side Transfer Network (STN) aims to fill the gap and capture uncertainty through granularity modeling. Intermediate features of SAM are gradually fused and projected into side outputs of edges at increasingly complex levels, allowing the model to fully and explicitly explore diverse uncertainties with granularity.

Specifically, we feed the image $X \in \mathbb{R}^{H \times W \times 3}$ into the frozen SAM with the prompt of $8 \times 8$ points grid, obtaining shallow feature maps $E_s$ (i.e. the output of the first block in the encoder), image embeddings $E_i$ from the SAM encoder, and mask embeddings $E_m$ from the decoder. 

For features $\{E_{s}, E_{i}\}$, we transform them into edge-aware features $\{E_{s}^{e}, E_{i}^{e}\} \in \mathbb{R}^{D \times D \times C_1}$ using learnable convolution layers, where $D \times D$ represents the spatial resolution and $C_1$ is feature channels. For the mask embedding feature $E_{m}$, we first rescale the feature channel to $C_2$ using learnable convolutional layers, and then reshape the features into edge-aware features $E_{m}^{e} \in \mathbb{R}^{D \times D \times (C_2 \times (8 \times 8))}$, ensuring all edge-aware details from the prompt is integrated. The detailed process is formulated as follows:
\begin{equation}
\begin{aligned}
    E_i^e = W_i^t(E_i), \, E_s^e = W_s^t(E_s), \, E_m^e = R(W_m^t(E_m))\\
\end{aligned}
\label{eq:pipline_pretransfer}
\end{equation}
where $W_i^t(\cdot),W_s^t(\cdot),W_m^t(\cdot)$ are convolution layers, $R(\cdot)$ is the reshape operator.

The Feature Fusion Block (FFB) is developed inside the STN to gradually aggregate edge-aware features $\{E_{s}^{e}, E_{i}^{e}, E_{m}^{e}\}$. Inspired by ~\cite{wu2023learning}, we construct the FFB as a stack of cross-attention operations and Gated-Dconv Feed-Forward Network proposed by ~\cite{zamir2022restormer}. The process of constructing edge features at different granularity levels can be summarized as follows:
\begin{equation}
\begin{aligned}
    F^c &= W_c^h(E_i^e), E^m = \text{FFB}_{1}(E_i^e \cdot W_c^g(F_c), E_s^e),\\
    F^m &= W_m^h(E^m), E^f = E^m \cdot W_m^g(F_m), \\
    F^f &= W_f^h({E^f}^{\prime}), {E^f}^{\prime} = \text{FFB}_{2}(E^f, E_m^e),
\end{aligned}
\label{eq:pipline_features}
\end{equation}
where $W_c^h(\cdot),W_m^h(\cdot),W_f^h(\cdot), W_c^g(\cdot), W_m^g(\cdot)$ are convolution layers, $\text{FFB}(\cdot)$ is the Feature Fuse Block. Specifically, we directly generate the coarse-grained edge feature $F^c$ using the $E_i^e$. Then, we use the first FFB to fuse $E_i^e$ and $E_s^e$ for including more details, which forms our medium-grained edge feature $F_m$. 
The features are further fused with $E_m^e$ by another FFB, resulting our fine-grained edge feature $F_f$.

\begin{figure*}[ht]
	\centering
	\includegraphics[width= 0.9 \linewidth, height = 0.25\linewidth]{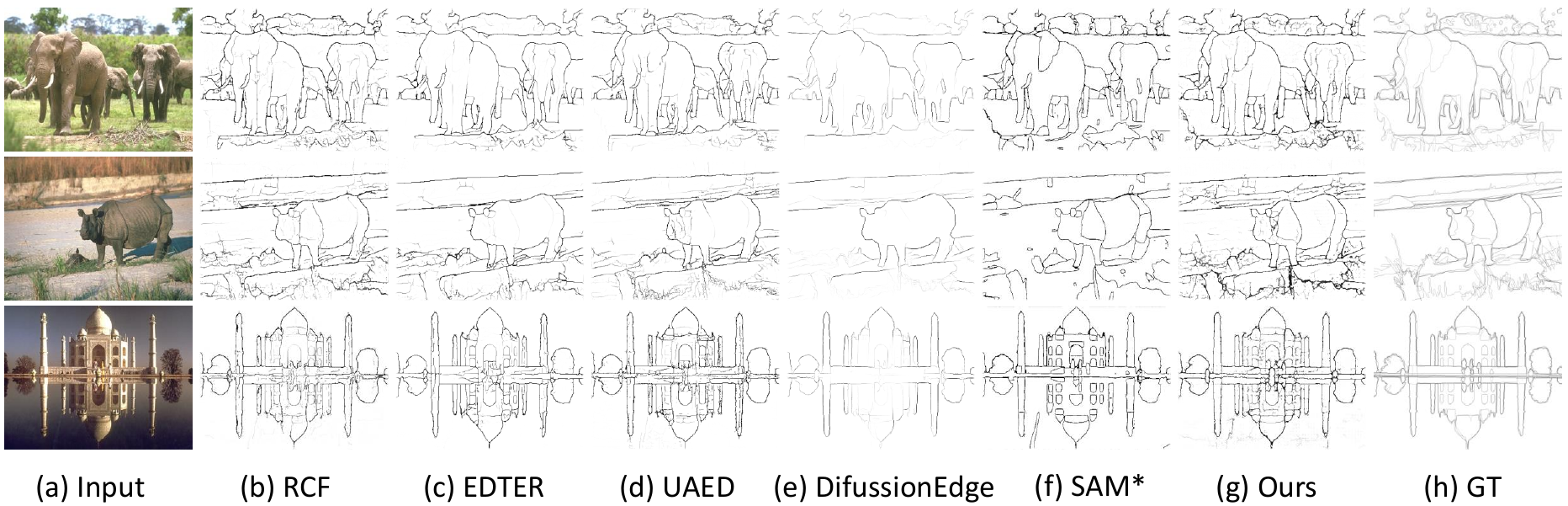}
	\caption{Qualitative comparison results on BSDS test set. * indicates the Zero Shot method.}
	\label{fig:qualitative_bsds}
\end{figure*}

\begin{figure}[ht]
	\centering
	\includegraphics[width= 0.91\linewidth, height = 0.63\linewidth]{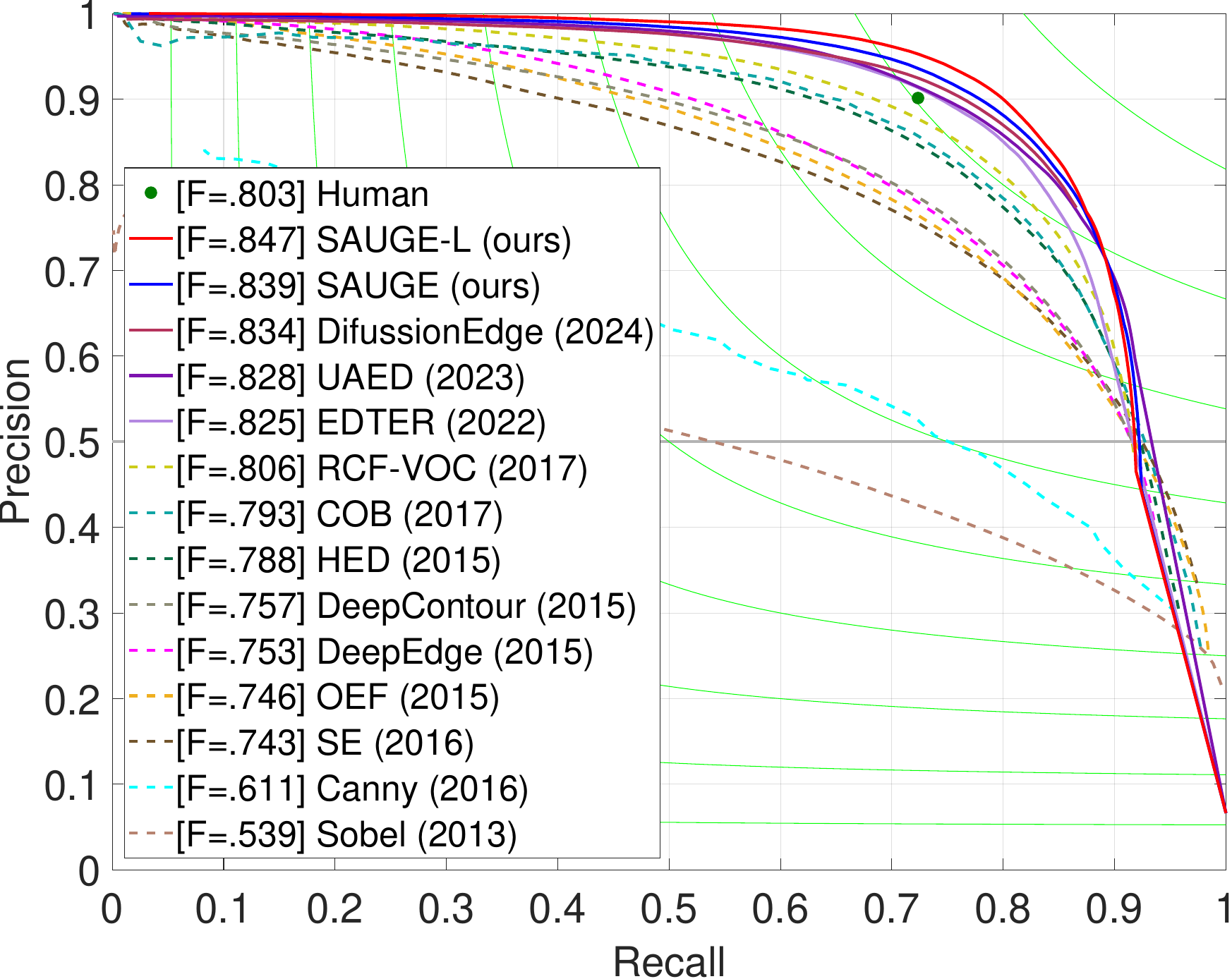}
	\caption{Precision-recall curves for the BSDS500 test set.}
	\label{fig:pr_curve}
\end{figure}

With the edge features $\{F^c,F^m,F^f\}$ prepared, we then employ a shared classification head $H$ to generate the side outputs $\{\hat{Y}^c, \hat{Y}^m, \hat{Y}^f\}$, which represents edges at different granularity levels. The features of these side outputs are then concatenated to form the final output $\hat{Y}^u$. The detailed process is elaborated as follows:
\begin{equation}
\begin{aligned}
    \hat{Y}^c &= H(F^c), \Ddot{F}^m = W_m^a([F^m,F^c]), \\
    \hat{Y}^m &= H(\Ddot{F}^m), \Ddot{F}^f = W_f^a([F^f,\Ddot{F}^m]),\\
    \hat{Y}^f &= H(\Ddot{F}^f), \hat{Y}^u  = H(W_u^a([F^c, \Ddot{F}^m, \Ddot{F}^f])),
\end{aligned}
\label{eq:pipline}
\end{equation}
where $W_m^a(\cdot),W_f^a(\cdot),W_u^a(\cdot)$ are convolutions, $H$ is output head shared across varied side outputs, $[\cdot]$ is concatenation.

\subsection{Generation of edge at arbitrary granularity} \label{sec:mg}
Given side outputs at three granularity levels (coarse-grained $\hat{Y}^c$, medium-grained $\hat{Y}^m$, fine-grained $\hat{Y}^f$), we propose to generate edge maps of arbitrary granularity $\hat{Y}^\alpha$ by a simple linear weighting strategy. Specifically, following the work of MuGE~\cite{zhou2024muge}, we measure the granularity level using $\alpha \in [0,1]$, where $0$ represents the coarsest and $1$ represents the finest. 
Then, the process of generating arbitrary granularity edge graph $\hat{Y}^\alpha$ can be formulated as follows:
\begin{equation}
\hat{Y}^\alpha=
\begin{cases}
    \frac{\alpha}{0.5}\hat{Y}^m + \left(1 - \frac{\alpha}{0.5}\right)\hat{Y}^c, & 0 \leq \alpha \leq 0.5\\
    \frac{\alpha - 0.5}{0.5}\hat{Y}^f + \left(1 - \frac{\alpha - 0.5}{0.5}\right)\hat{Y}^m, & 0.5 < \alpha \leq 1
\end{cases}
\label{eq:any-granularity}
\end{equation}

\subsection{Loss functions for model training}
Three loss functions are formulated to train our framework with three side outputs and one final output for detecting edges at multiple granularity: the granularity-aligned loss function for supervising side outputs with pseudo labels, the diversity loss function for enhancing the difference between different side outputs, and the guide loss constraints the produced edges to be compatible with SAM masks.

\textbf{Pseudo labels at varied granularities.}
Given the input image $X$ and its manual annotations $\{{Y _ n}\}^ {N}_ {n=1}$, where $N$ is the number of annotations, and $Y _ n \in \{{0,1}\} ^ {H \times W}$ is the annotation given by the $n$-th annotator.  
We sort all annotations in ascending order based on the number of pixels labeled as edges. Denote the sorted annotations as $\{{Y_{n} ^{S}\}}^ {N}_ {n=1}$, then the coarse-grained annotation $Y^c$, medium grained annotation $Y^m$, and fine-grained annotation $Y^f$ can be constructed as:
    \begin{equation}
    \hspace*{\fill}
    Y^c = Y_1^S, \hfill Y^m = Y^c \lor Y_{\lceil \frac{N}{2} \rceil}^{S}, \hfill Y^f = Y^m \lor Y_N^S,
    \hspace*{\fill}
    \label{eq:label}
    \end{equation}
where $\lor$ is element-wise OR operation. The side outputs at different granularity levels $\{\hat{Y}^c, \hat{Y}^m, \hat{Y}^f\}$ are supervised by the corresponding $\{Y^c, Y^m, Y^f\}$. 

In order to comprehensively consider all annotations and catch the uncertainty caused by multi-label, inspired by ~\cite{zhou2023treasure}, the distribution of final label $\tilde{Y}^u$ is obtained by randomly sampling from a multivariate Gaussian distribution $N (\mu_Y, \sigma_Y)$ and then binarizing $\tilde{Y}^u$ to the final label $Y^u$ using the threshold $\zeta$, where the mean $\mu_Y$ and variance $\sigma_Y$ are pixel-wise calculated by the label set $\{{Y _ n}\}^ {N}_ {n=1}$.

\textbf{Granularity-aligned loss for side outputs.}
Edge detection is a binary classification task where each pixel needs to be predicted as an edge (positive) or not (negative), hence binary cross-entropy(BCE) loss is widely used in this task. However, the edge pixels in an image are usually only a tiny fraction of all pixels. Following ~\cite{xie2015holistically}, we adaptively weight each pixel based on the ratio of positive and negative samples in the ground truth. For the side outputs $\{\hat{Y}^c, \hat{Y}^m, \hat{Y}^f\}$, we minimize the following BCE loss:
\begin{equation}
\begin{aligned}
    L_{detect}^{g} & = - \sum_{j=1}^{HW}(Y_j^{g} \xi \text{log}(\hat{Y}_j^{g})) \\
    & + (1 - Y_j^{g})(1 - \xi) \text{log}(1 - \hat{Y}_j^{g})),
\end{aligned}
\label{eq:side_loss}
\end{equation}
where $\xi$ is used to balance the contribution of varied pixels, $g \in \{c,m,f,u\}$ represents the granularity level, and $j$ represents the $j$-th pixel in the prediction and ground truth. In addition, $\xi = \lvert Y_{-}^{g} \rvert / (\lvert Y_{-}^{g} \rvert + \lvert Y_{+}^{g} \rvert)$, $\lvert \cdot \rvert$ represents the total number of pixels, and $Y_{+}^{g}$ and $Y_{-}^{g}$ represents positive and negative samples in ground truth $Y^g$.

Given the pseudo labels at varied granularities $\{Y^c, Y^m, Y^f\}$, the loss for supervising the side outputs to align granularity is defined as:
\begin{equation}
\begin{aligned}
    L_{side} = \sum_{g \in \{c,m,f\}}L_{detect}^{g},
\end{aligned}
\label{eq:side_loss_total}
\end{equation}

\begin{table}[t]
\small
\setlength{\tabcolsep}{1mm}
\centering
\begin{tabular}{c|c|c|ccc}
\hline
& Method & Param. (M) & ODS & OIS & AP \\ \hline
\multirow{19}{*}{\begin{tabular}[c]{@{}c@{}} \rotatebox{90}{Single-Scale (SS)}\end{tabular}} 
& Canny \hfill{\small{(PAMI'86)}} & - & .611 & .676 & .520 \\ 
& OEF \hfill{(CVPR'15)} & - & .746 & .770 & .815 \\
& DeepContour \hfill{(CVPR'15)} & 27.5 & .757 & .776 & .790 \\ 
& DeepBoundary \hfill{(ICLR'15)} & - & .789 & .811 & .789 \\
& HED \hfill{(ICCV'15)} & - & .788 & .808 & .840 \\
& RDS \hfill{(CVPR'16)} & - & .792 & .810 & .818 \\ 
& RCF \hfill{(CVPR'17)} & 14.8 & .798 & .815 & - \\
& CED \hfill{(CVPR'17)} & 21.8 & .803 & .820 & .871 \\
& BDCN \hfill{(CVPR'19)} & 16.3 & .806 & .826 & .847 \\
& DSCD \hfill{(MM'20)} & - & .802 & .817 & - \\
& LDC \hfill{(MM'21)} & - & .799 & .816 & .837 \\
& EDTER \hfill{(CVPR'22)} & 468.8 & .824 & .841 & .880 \\
& UAED \hfill{(CVPR'23)} & 69.2 & .829 & .847 & .892 \\
& RankED \hfill{(CVPR'24)} & 114.4 & .824 & .840 & \underline{.895} \\
& DiffusionEdge \hfill{(AAAI'24)} & 224.9 & .834 & .848 & .815 \\
& EdgeSAM \hfill{(TII'24)} & 7.4 & .838 & .852 & .893 \\
& SAUGE \hfill{(Ours)} & \textbf{1.3} & \underline{.839} & \underline{.860} & .893 \\
& SAUGE-L \hfill{(Ours)} & \underline{2.1} & \textbf{.847} & \textbf{.868} & \textbf{.898} \\
\hline
\multirow{11}{*}{\begin{tabular}[c]{@{}c@{}} \rotatebox{90}{Single-Scale-VOC (SS-VOC)}\end{tabular}} 
& DeepBoundary \hfill{(ICLR'15)} & - & .809 & .827 & .861 \\
& RCF \hfill{(CVPR'17)} & 14.8 & .806 & .823 & - \\
& CED \hfill{(CVPR'17)} & 21.8 & .812 & .833 & .889 \\
& BDCN \hfill{(CVPR'19)} & 16.3 & .820 & .838 & .888 \\
& DSCD \hfill{(MM'20)} & - & .813 & .836 & - \\
& LDC \hfill{(MM'21)} & - & .812 & .826 & .857 \\
& EDTER \hfill{(CVPR'22)} & 468.8 & .832 & .847 & .886 \\
& UAED \hfill{(CVPR'23)} & 69.2 & .838 & .855 & \textbf{.902} \\
& RANKED \hfill{(CVPR'24)} & 114.4 & .833 & .848 & \underline{.901} \\
& SAUGE \hfill{(Ours)} & \textbf{1.3} & \underline{.842} & \underline{.862} & .896 \\
& SAUGE-L \hfill{(Ours)} & \underline{2.1} & \textbf{.849} & \textbf{.869} & .899 \\
\hline
\end{tabular}
\caption{Comparisons of approaches with single edge output on BSDS500 test set.  The best and second-best results are shown with bold and underlined texts.}
\label{tab:bsds_comparison}
\end{table}

\textbf{Emphasize the difference between side outputs.}
Different side outputs represent edge maps at different granularity levels, we use mean absolute error (MAE) to emphasize the diversity of side outputs:
\begin{equation}
\begin{aligned}
    L_{differ}^{(i,k)} &= - \sum_{j=1}^{HW}(\lvert \hat{Y}_j^{i} - \hat{Y}_j^{k} \rvert) \cdot ( Y_j^{i} \oplus Y_j^{k}),
\end{aligned}
\label{eq:differ_loss}
\end{equation}
where $(i, k)$ is about the combination of $\{c, m, f\}$ with a size of $2$, $\lvert \cdot \rvert$ represents taking the absolute value, $\oplus$ represents element-wise XOR operation. In this way, different pixels with edge definitions can be located, and two different side outputs can have differences on these pixels. $L_{differ}$ is obtained by enumerating combinations $(i, k)$ and summing $L_{differ}^{(i,k)}$ as:
\begin{equation}
\begin{aligned}
    L_{differ}= L_{differ}^{(c,m)} + L_{differ}^{(c,f)} + L_{differ}^{(m,f)},
\end{aligned}
\label{eq:total_differ_loss}
\end{equation}
The use of $L_{differ}$ promotes the generation of rougher $\hat{Y}^c$ (Figure \ref{fig:side_differ}), ensuring the difference between the side outputs.

\textbf{Guide loss for final output.}
 Here, we further use the object masks outputted by SAM to guide our edge learning. Given the object mask $M$ outputted by SAM, we use the Sobel~\cite{kittler1983accuracy} operator to extract the edges of each mask and denote it as $Y^{mask}$. Meanwhile, we calculate maps $\overline{Y}^{mask} \in \mathbb{R}^{H \times W}$ which indicate the frequency of the corresponding pixel is recognized as edge on each mask.

The loss for supervising the final output $\hat{Y}^u$ is defined as:
\begin{equation}
\begin{aligned}
    L_{guide} = \sum_{j=1}^{HW}e^{(\psi_j + \omega_j)} L_{detect}^u
\end{aligned}
\label{eq:guide_loss}
\end{equation}
where $\psi_j = - (Y^{mask}_j) \cdot (Y^{mask}_j \oplus Y_j) \cdot \overline{Y}^{mask}_j$, $\omega_j = cos(\tilde{Y}^u_j)$. The weight map $\psi$ is defined in the way such that the easily confused pixels indicated by the object mask contributes less in the edge learning. $\omega$ is constructed in order to assign higher weights to edge pixels with lower confidence.

\textbf{Overall loss function.}
    The final optimization objective is defined as the weighted sum of $L_{side}$, $L_{differ}$, and $L_{guide}$:
\begin{equation}
\begin{aligned}
    L_{total} = L_{guide} + \lambda L_{differ} + \beta L_{side}
\end{aligned}
\label{eq:overall_loss}
\end{equation}
where $\lambda$ and $\beta$ are the coefficient to control the balance of different losses. In this work, we fix $\lambda = 0.1, \beta = 0.5$.

\begin{table}[t]
\small
\setlength{\tabcolsep}{1mm}
\centering
\begin{tabular}{c|c|c|ccc}
\hline
& Method & Param. (M) & ODS & OIS & AP \\ \hline
\multirow{4}{*}{\begin{tabular}[c]{@{}c@{}} \rotatebox{90}{SS}\end{tabular}} 
& MuGE (M=3) \hfill{(CVPR'24)} & 93.9  & .845 & .854 & .876 \\
& SAUGE (M=3) \hfill{(Ours)} & 1.3 & \underline{.854} & \textbf{.865} & \underline{.894} \\
& MuGE (M=11) \hfill{(CVPR'24)} & 93.9 & .850 & .856 & .882 \\
& SAUGE (M=11) \hfill{(Ours)} & 1.3 & \textbf{.857} & \textbf{.865} & 
\textbf{.896} \\
\hline
\multirow{4}{*}{\begin{tabular}[c]{@{}c@{}} \rotatebox{90}{SS-VOC}\end{tabular}} 
& MuGE (M=3) \hfill{(CVPR'24)} & 93.9 & .852 & .859 & .889 \\
& SAUGE (M=3) \hfill{(Ours)} & 1.3 & \underline{.857} & \underline{.867} & \underline{.900} \\
& MuGE (M=11) \hfill{(CVPR'24)} & 93.9 & .855 & .860 & .894 \\
& SAUGE (M=11) \hfill{(Ours)} & 1.3 & \textbf{.859} & \textbf{.868} & 
\textbf{.901} \\
\hline
\end{tabular}
\caption{Comparison of multi-granularity approaches on the BSDS500 test set. The best and second-best results are shown with bold and underlined texts, respectively.}
\label{tab:bsds_mg_comparison}
\end{table}

\section{Experiments}
\subsection{Datasets}
We conduct experiments on three widely-used edge detection datasets: BSDS500~\cite{arbelaez2010contour},  Multicue~\cite{mely2016systematic} and NYUDv2~\cite{silberman2012indoor}. For data augmentation, we adopt the same strategy as UAED~\cite{zhou2023treasure} across both datasets.

\textbf{BSDS500} consists of 500 natural images, with 200 for training, 100 for validation, and the remaining for test. Each image has 4 to 9 manual annotations. 
Additionally, the PASCAL VOC set~\cite{everingham2010pascal} with 10,103 images is used as supplementary training data, with edge annotations derived from semantic masks using Laplacian detector.

\textbf{Multicue} includes 100 images from complex natural scenes, each of which is annotated by multiple individuals. We randomly split these images into training and evaluation sets, with 80 images for training and 20 for testing. This process is repeated three times and average scores are reported.

\textbf{NYUDv2} is a dataset for indoor scene parsing and edge detection, containing 1,449 paired RGB-D images. Each image has a single ground-truth edge map, with the dataset split into 381 training, 414 validation, and 654 testing images.

\begin{table}[t]
\small
\centering
\begin{tabular}{c|ccc}
\hline
Methods & ODS & OIS & AP \\
\hline
Human \hfill{\small{(VR'16)}} & 0.750 & - & - \\
Multicue \hfill{\small{(VR'16)}} & 0.830 & - & - \\
HED \hfill{\small{(ICCV'15)}} & 0.851 & 0.864 & - \\
RCF \hfill{\small{(CVPR'17)}} & 0.857 & 0.862 & - \\
BDCN \hfill{\small{(CVPR'19)}} & 0.891 & 0.898 & 0.835 \\
DSCD \hfill{\small{(MM'20)}} & 0.871 & 0.876 & - \\
LDC \hfill{\small{(MM'21)}} & 0.881 & 0.893 & - \\
EDTER \hfill{\small{(CVPR'22)}} & 0.894 & 0.900 & 0.944 \\
UAED \hfill{\small{(CVPR'23)}} & 0.895 & 0.902 & \underline{0.949} \\
MuGE \hfill{\small{(CVPR'24)}} & 0.898 & 0.900 & \textbf{0.950} \\
DifussionEdge \hfill{\small{(AAAI'24)}} & \underline{0.904} & \textbf{0.909} & - \\
SAUGE \hfill{\small{(Ours)}} & \textbf{0.905} & \underline{0.907} & 0.939 \\
\hline
\end{tabular}
\label{tab:multicue_comparison}
\caption{Comparisons on Multicue. Best and second-best results are shown with bold and underlined texts, respectively. 
}
\end{table}

\subsection{Implementation Details}
We implement SAUGE based on PyTorch~\cite{paszke2019pytorch}, and use SAM pre-trained on the SA-1B dataset~\cite{kirillov2023segment} as our backbone. The Adam optimizer~\cite{kinga2015method} is used to update all parameters. The learning rate is initialized as 1e-4 with step scheduling and weight decay is set to 5e-4. 
For BSDS, we set the $\zeta$ for the thresholding label to $0.2$, and the model was trained for 6 epochs with a batch size of 3. For Multicue, we randomly crop the images into $512 \times 512$ and set $\zeta$ to $0.3$. The model is trained for 20 epochs using a batch size of 3.
All experiments were conducted on RTX 3090, where training the model on BSDS500 requires approximately 20 GPU hours and 16GB GPU memory.

\subsection{Evaluation Metric}
We evaluate the performance using widely adopted metrics, including F-scores for both Optimal Dataset Scale (ODS) and Optimal Image Scale (OIS), as well as Average Precision (AP). 
ODS applies a fixed threshold to binarize the predicted edge map across the entire dataset, while OIS selects an optimal threshold for each image. We also report the learnable parameter. Following previous works~\cite{liu2017richer,pu2022edter,zhou2023treasure}, we apply non-maximum suppression (NMS) to the predictions before evaluation. The localization tolerance is set to 0.011 for NYUDv2 and 0.0075 for other datasets, defining the maximum allowed distance for matching predictions with GT. For AP evaluation, we select the candidate with the highest F-value across M candidates at each threshold, providing a more comprehensive assessment of the overall performance across all candidates. To assess the quality of multiple granularity edge maps (M=3, 11, etc), we follow the MuGE~\cite{zhou2024muge} that uses the best-matching strategy, selecting the candidate closest to the GT from multiple candidates.

\subsection{Comparison with State-of-the-arts}
Existing methods can be divided into two categories: single definite output and multi-granularity output. For single definite output, 
traditional methods include Canny~\cite{canny1986computational} and OEF~\cite{hallman2015oriented}. CNN-based methods, such as DeepContour~\cite{shen2015deepcontour}, DeepBoundary~\cite{kokkinos2015pushing}, HED~\cite{xie2015holistically}, RDS~\cite{liu2016learning}, CED~\cite{wang2017deep}, BDCN~\cite{he2019bi}, DSCD~\cite{deng2020deep}, and LDC~\cite{deng2021learning}, are widely adopted. Transformer-based methods involve EDTER~\cite{pu2022edter}, RankED ~\cite{cetinkaya2024ranked} and EdgeSAM~\cite{yang2024boosting}.
The diffusion-based method involves DiffusionEdge~\cite{ye2024diffusionedge}.
Currently, only MuGE~\cite{zhou2024muge} is developed for multi-granularity output.

\begin{table}[t]
\small
\centering
\begin{tabular}{c|ccc}
\hline
Methods & ODS & OIS & AP \\
\hline
OEF \hfill{\small{(CVPR'15)}} & 0.651 & 0.667 & - \\
HED \hfill{\small{(ICCV'16)}} & 0.720 & 0.734 & 0.734 \\
RCF \hfill{\small{(CVPR'17)}} & 0.729 & 0.742 & - \\
BDCN \hfill{\small{(CVPR'19)}} & 0.748 & 0.763 & 0.770 \\
EDTER \hfill{\small{(CVPR'22)}} & 0.774 & 0.789 & 0.797 \\
SAM* \hfill{\small{(ICCV'23)}} & 0.699 & 0.719 & 0.707 \\
DiffusionEdge \hfill{\small{(AAAI'24)}} & 0.761 & 0.766 & - \\
EdgeSAM \hfill{\small{(TII'24)}} & \underline{0.783} & \underline{0.797} & 0.805\\
RankED \hfill{\small{(CVPR'24)}} & 0.780 & 0.793 & \textbf{0.826} \\
SAUGE-L* \hfill{\small{(Ours)}} & \textbf{0.794} & \textbf{0.803} & \underline{0.813} \\
\hline
\end{tabular}
\caption{Comparisons on the NYUDv2. The best and second-best results are shown with bold and underlined texts respectively. * indicates the Zero Shot method.}
\label{tab:nyud_comparison}
\end{table}

\begin{figure*}[ht]
	\centering
	\includegraphics[width=\linewidth, height=0.15\linewidth]{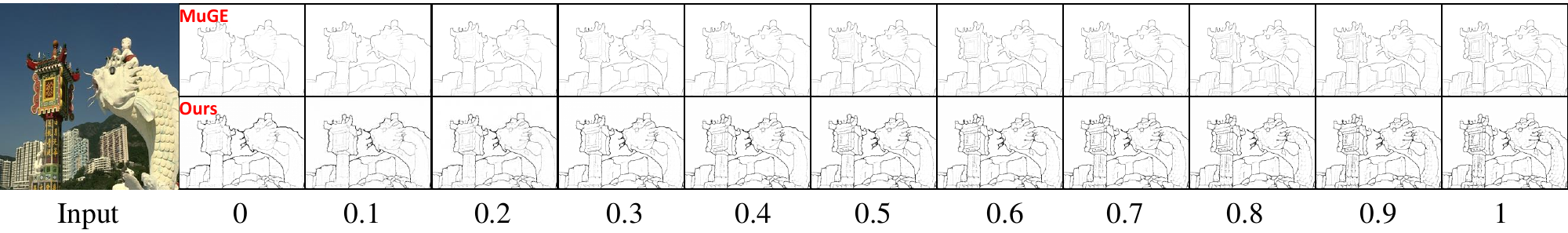}
	\caption{Qualitative comparison with MuGE for different edge granularities $\alpha$ under the SS-VOC setting.}
	\label{fig:multi-granularity}
\end{figure*}

\begin{figure}[htbp]
	\centering
	\includegraphics[width=\linewidth, height=0.445\linewidth]{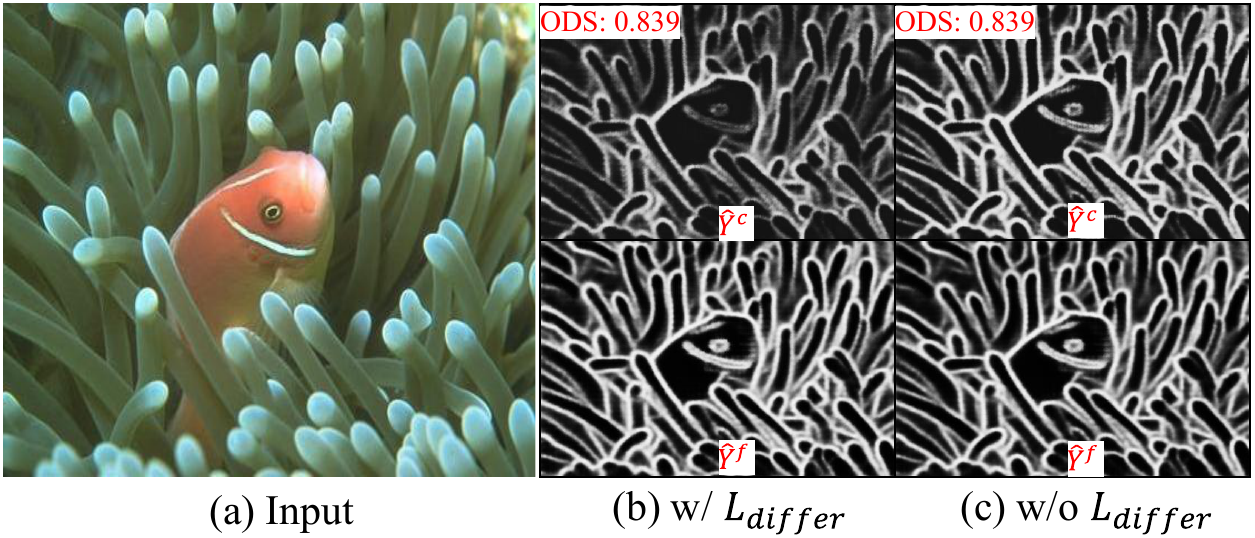}
	\caption{Comparison of coarse-grained and fine-grained side output $\{\hat{Y}^c, \hat{Y}^f\}$ with/without $L_{differ}$. The ODS is obtained on BSDS500 test set under Single-Scale (SS) setting.}
	\label{fig:side_differ}
\end{figure}

\textbf{Results on BSDS500.} 
Since SAUGE can generate both definite output $\hat{Y}^u$ and multi-granularity outputs $\hat{Y}^\alpha$, we compare it with excellent representatives of these two types of methods respectively. The quantitative results are presented in Table \ref{tab:bsds_comparison} and Table \ref{tab:bsds_mg_comparison}, where $M$=$3$ represents $\alpha$=$\{0,$ $0.5,1\}$, and $M$=$11$ represents the 11 outputs generated from 0 to 1 at intervals of 0.1. Experiments were conducted under two settings: Single-Scale input (SS) and Single-Scale input with additional  PASCAL VOC data for training (SS-VOC).

In Table \ref{tab:bsds_comparison}, we report the performance of existing state-of-the-art methods for definite output. As shown, our SAUGE achieves the best performances in both ODS and OIS metrics under all experimental settings (SS and SS-VOC), which significantly outperforms EdgeSAM with the same SAM backbone.  It is worth noting that our model only needs to tune a very small number of parameters—82\% fewer compared to EdgeSAM. When using a larger SAM (ViT-L SAM) as the backbone (denoted as SAUGE-L), the performance improves by 0.009 in ODS and 0.016 in OIS under the SS setting, and by 0.011 in ODS and 0.014 in OIS under the SS-VOC setting. Qualitatively, Figure \ref{fig:qualitative_bsds} compares our SAUGE with other methods, demonstrating that SAUGE can produce more detailed predictions. We note that the background edges missed by SAM can be successfully obtained by our model. The Precision-Recall curve is shown in Figure \ref{fig:pr_curve}. 

For multi-granularity output, Table \ref{tab:bsds_mg_comparison} shows that SAUGE significantly outperforms MuGE in all settings. It can be found that increasing the number of candidates (M) leads to higher performance, demonstrating the advantages of multi-granularity edge modeling in addressing uncertainty.

\textbf{Results on Multicue.}
We also conduct experiments on Multicue edges, and the results are illustrated in Table \ref{tab:multicue_comparison}. As can be seen, the proposed SAUGE outperforms the previous state-of-the-art in the ODS metric and achieves second-place performance in the OIS metric, with scores of 90.5\% in ODS and 90.7\% in the OIS metric. The relatively lower performance in the AP metric may be attributed to overfitting, likely due to the limited size of the Multicue dataset.

\textbf{Results on NYUDv2.}
To verify the generalization of our SAUGE, we directly use the SAUGE-L model trained on BSDS500 and PASCAL VOC for zero-shot edge detection on this set. The results in Table \ref{tab:nyud_comparison} indicate that our method outperforms previous supervised methods on both ODS and OIS without parameter tuning, which demonstrates the effectiveness of our approach.

\subsection{Ablation Study and Further Analysis}
\textbf{Effect of key components.} 
The key designs of our SAUGE lie in the lightweight Side Transfer Network (STN), the constraints on the multi-granularity side outputs, and the mask-guided loss $L_{guide}$. To verify the effectiveness of these designs, we begin with SAM as the baseline, and then sequentially add the Side Transfer Network, the side output constraints (denoted as SOC), and the mask-guided loss. 
The results on the BSDS500 dataset are presented in Table \ref{tab:method_ablation}.
As shown, for both pre-trained ViT-B SAM and ViT-L SAM baseline, the inclusion of our lightweight STN module leads to a significant improvement across all metrics, by more than 9\% in ODS, 8.7\% in OIS, 10.8\% in AP metrics. Also, with the constraints on side output (SOC), our framework can generate edges at arbitrary granularity and achieve a notable enhancement in the AP metric. We attribute this to the fact that SOC explicitly facilitates the model in capturing different granularity information, thus enabling better alignment with the uncertainty in edge decisions. Additionally, mask-guided loss $L_{guide}$ also contributes positively to all metrics. To study the impact of $L_{differ}$, Figure \ref{fig:side_differ} visualizes model outputs with and without it. As shown, the $L_{differ}$ amplifies the differences between outputs at various granularities, further enhancing alignment with edge decision uncertainty.

\begin{table}[t]
\centering
\small
\setlength{\tabcolsep}{2.5mm}
\begin{tabular}{c|cc|cccc}
\hline
Method & SOC &$\mathcal{L}_{guide}$ & ODS & OIS & AP & MG\\ \hline
\multirow{1}{*}{SAM*}  &  &  & .768 & .786 & .794 & $\times$   \\ \hline
\multirow{3}{*}{Ours} 
& & & .837 & .855 & .880 & $\times$ \\ 
& \checkmark &  & .836 & .857 & .891 & $\checkmark$\\
& \checkmark & \checkmark & .839 & .860 & .893 & $\checkmark$\\
\hline
\multirow{3}{*}{Ours-L}
&  &  & .844 & .865 & .822 & $\times$   \\
& \checkmark &  & .846 & .866 & .895 & $\checkmark$\\
& \checkmark & \checkmark & .847 & .868 & .898 & $\checkmark$\\
\hline
\end{tabular}
\caption{Ablation study on key components of our model on the BSDS500 set. MG refers to whether the method can output multi-granularity edges. * indicates Zero Shot method.}
\label{tab:method_ablation}
\end{table}

\begin{table}[t]
\small
\setlength{\tabcolsep}{1.5mm}
\centering
\begin{tabular}{c|ccccc}
\hline
Performance & $\alpha=0$ & $\alpha=0.2$ & $\alpha=0.4$ & $\alpha=0.5$ & $\alpha=1$ \\ \hline
ODS & 0.832 & 0.836 & 0.839 & 0.839 & 0.836\\
OIS  & 0.851 & 0.857 &0.859 & 0.859 & 0.858\\
AP & 0.771 & 0.814 &0.882 & 0.890 & 0.891\\
\hline
\end{tabular}
\caption{Comparisons on BSDS500 for varied granularities.}
\label{tab:single_comparison}
\end{table}

\textbf{Performance of specific granularity outputs.} 
Given a granularity level $\alpha$, our method can generate edge maps at arbitrary granularity in a controlled manner. In Figure \ref{fig:multi-granularity}, we visualize the edge maps at different granularities generated by MuGE and our SAUGE. For all $\alpha$, SAUGE consistently outperforms MuGE and produces edge maps with high quality.
Furthermore, we report the quantitative results of various granularity $\alpha$=\{0, 0.2, 0.4, 0.5, 1\} in Table \ref{tab:single_comparison}. As shown, the edge maps at varied granularities exhibit both good performance and diversity in ODS, OIS and AP metrics.

\textbf{Computational efficiency.}
We evaluate the inference efficiency of our proposed SAUGE on a single RTX 3090 GPU. SAUGE consumes 6.5GB of GPU memory and achieves a speed of 5.3 FPS, exhibiting similar efficiency to the transformer-based EDTER (6.0GB, 5.3 FPS) while outperforming the zero-shot-based SAM (8.1GB, 0.9 FPS) and the diffusion-based DiffusionEdge (3.4GB, 0.4 FPS) in speed.

\section{Conclusion and Limitation}
In this paper, we designed a novel uncertainty-aligned edge detector called SAUGE. In our SAUGE, we developed a lightweight Side Transfer Network (STN) to explicitly explore the knowledge embedded in SAM for multiple granularity edge detection. Extensive experiments on three edge sets are reported to demonstrate the superiority of SAUGE.

\paragraph{Limitation.}
Our SAUGE is constructed based on the SAM backbone model. In the future, we would like to extend our framework for a more efficient backbone model.

\section{Acknowledgments}
This work was supported partially by the NSFC (U21A20- 471, U22A2095, 62076260, 61772570), Guangdong Natural Science Funds Project (2023B1515040025), Guangdong NSF for Distinguished Young Scholar (2022B15- 15020009), Guangdong Provincial Key Laboratory of Information Security Technology (2023B1212060026),  open research fund of Key Laboratory of Machine Intelligence and System Control, Ministry of Education (No. MISC-202407). 

\bibliography{aaai25}

\end{document}